# Intelligent requirements engineering from natural language and their chaining toward CAD models

**Alain-Jérôme FOUGÈRES**
ECAM Rennes
Campus de Ker Lann – Bruz, 35091 Rennes, France
alain-jerome.fougeres@ecam-rennes.fr
**Egon OSTROSI**
Université de Bourgogne Franche-Comté, UTBM
ERCOS/ELLIADD EA4661, Belfort, France
egon.ostrosi@utbm.fr

## ABSTRACT

*This paper assumes that design language plays an important role in how designers design and on the creativity of designers. Designers use and develop models as an aid to thinking, a focus for discussion and decision-making and a means of evaluating the reliability of the proposals. This paper proposes an intelligent method for requirements engineering from natural language and their chaining toward CAD models. The transition from linguistic analysis to the representation of engineering requirements consists of the translation of the syntactic structure into semantic form represented by conceptual graphs. Based on the isomorphism between conceptual graphs and predicate logic, a formal language of the specification is proposed. The outcome of this language is chained and translated in Computer Aided Three-Dimensional Interactive Application (CATIA) models. The tool (EGEON: EngineerinG desiGn sEmantics elabOration and applicatioN) is developed to represent the semantic network of engineering requirements. A case study on the design of a car door hinge is presented to illustrates the proposed method.*

## 1.  INTRODUCTION

Designers make use of a variety of forms of knowing as they move from conceptualization to representation. Linguistic modelling occurs when what they are envisaging or cognitively modelling is presented in words and designers must, of course, employ the words of natural language when they are   modelling their designs linguistically.

Much of designing is, in fact, the making of meaning. Words have definitions but, at the same time, have no clear-cut boundaries to their meanings. People might, of course, when asked to explain the meaning of a word, mean the explanation which, on being asked, they are ready to give [1], that is, if they are ready to give any explanation, which, in many cases, they are not. Many words do not have strict meanings. Therefore, one designer's clear meaning can be another's fuzzy focus, even if they have an apparently shared experience. As the French mathematician Thom has proposed [2, 3], in principle, it can be possible to achieve hard conceptual boundaries in a mathematical language (geometrical and topological representation). But, in natural language, the capacity for many possible meanings is intrinsic, and the unfolding form has fuzzy boundaries and many possibilities for meaning.

When cognitive modelling is externalized in an engineering object (i.e., a drawing, schema, plans, *3D* model, or prototype), it demonstrates the capacity of a designer for engineering design modelling. In other words, it is one of the languages used by designers in developing their designs. Design language shapes how designers design and the expression of their creativity. Designers use and develop models as an aid to thinking, a focus for discussion and decision-making and a means of evaluating the reliability of their proposals.

Assisting *CAD* with natural language means expanding this process through asking a set of questions, such as: How to participate in the formalization process from natural language specifications of a product? How to extract the linguistic information contained in the definition of product design specifications which make it possible to elaborate formal or diagrammatic models? How to formalize a language of specification suitable for *CAD* modelling? These issues are addressed by the research in requirement engineering supported by Natural Language Processing (*NLP*) [4]. Software requirement engineering supported by *NLP* is a research theme developed over the last thirty years, the reader will find different approaches in [5, 6, 7, 8, 9, 10, 11, 12, 13, 14, 15, 16]. Much less research has been done on the application of *NLP* to design specification. As recent works we can cite: an approach on the ease of use or improved the design efficiency of *CAD/CAM* systems [17]; the use of *NLP* techniques to automatically generate parametric property-based requirements from unstructured and semi-structured specifications [18]; a natural language processing approach to identify ambiguous terms between different domains, and rank them by ambiguity score [19]; the application of *NLP* to automate the implementation of changes within the construction models [20].

Design assistance, as well as the translation of technical texts into the form of semi-formal or formal diagrams, for example, can be achieved by a process of semantic analysis and representation of these texts, using previously acquired linguistic and conceptual knowledge (i.e., specific areas previously conceptualized). The intention here is to provide more of an assistance to conceptualization than an automatic design. For example, in the following, a resolutely symbolic approach is adopted: an approach guided by the hypothesis of representations with formalisms of *AI* by analogy with mental representations (for other approaches, such as neural network or deep learning technics, see [21, 22, 23]). To understand is to construct successive formal representations more or less dependent on a set of statements, which are, then, translated into computational treatments. The work relies on a set of characteristics of these design specifications (texts):

- For reasons of efficiency and comprehension by their users, these design texts are most often concise, precise, coherent and unambiguous; this reflects the willingness of writers to cooperate and be understood by their readers, while maintaining the relevance of their writings [24].

- These texts are essentially descriptions of contexts, situations, objects, actions, or events. To interpret them, it is possible to use linguistic universals, such as the distinctions between noun and verb, between objects, and between action and relation. For a propositional analysis of discourses, a minimal typology of three verbal categories is generally proposed: 1) the stative, 2) the factive, and 3) declarative verbs. As for the characterization of objects, five voices are distinguished: 1) the existential, 2) the situative, 3) the equative, 4) the descriptive, and 5) the subjective. Effective design assistance seeks, first, to identify and, then, to represent the structural, functional, and dynamic elements contained in the sentences of the requirements expression texts. Thus, grammars which make it possible to determine the category of a proposition (action, event, state) and the arguments of the predicate (agent, object, source, etc.) are well adapted for these treatments [25]

- The lexicon and the sentence [26] are the most significant levels for the linguistic processing of these short, specific, and unambiguous design texts; their automatic understanding is thus facilitated, notably by computing the meaning of sentences from words (compositionality) and by using a knowledge representation formalism derived from *AI*, allowing a partial representation of meanings.

- According to Vygotsky [27], speaking (the spoken word) corresponds to the translation of an internal language (the thought) to an external language (language). Thus, understanding proceeds in a symmetrical way. Asking an individual to describe an activity in which he is an expert is to ask him to translate into an external language what he thinks of this activity. To facilitate the understanding of this description, he will try to translate his thought directly into an abbreviated language, characterized by a simplified syntactic form and purely predicative judgments. Indeed, understanding is conceptualizing.

The primary aim of this paper is to propose an intelligent method for requirements engineering from natural language and its chaining toward *CAD* models. The rest of this paper is organized as follows: Section 2 presents a method for engineering specifications, consisting of seven phases; Section 3 illustrates the proposed method by presenting a case study on the design of a car door hinge; and Section 4 provides the conclusions as well as a discussion of the study's implications for future research.

## 2. MODELING AND METHODOLOGY

The design activity can initially be a verbalization or formulation activity that can be memorized in documents written in natural language. Is it then possible to exploit procedures that facilitate the writing of formal specifications of the different design elements? Any procedure facilitating the writing of design specifications (i.e., more or less standardized texts) contributes to the reduction of the time inherent in the realization of not only this step but also the following steps.

Examining the problem as stated, four strong assumptions can be posed. They guide a new approach to engineering specifications and allow the development of a design formalization system [25]:

- A1: it is not possible to directly translate an informal design specification into a formal design specification. This assumption is directly correlated to the problem of translating a natural language into a formal language. Moreover, the observation of co-design activity [28] corroborates this assumption; indeed, a conception (formal or not) is gradually elaborated (by refinement).

- A2: it is relevant to use an intermediary representation to move from the informal to the formal. The notion of refining the development of formal design specifications is not sufficiently defined to uniquely establish the number of refinement steps required for a given specification. Therefore, an intermediate representation or the pivot of the formalization is proposed. For instance, this intermediate representation might be in the form of conceptual graphs [29]. This proposition does not solve all the difficulties, but it allows better characterize of them: the complexity of the translation of a natural semantics into a formal semantics remains, but it doubles up and becomes easier to master.

- A3: the formal representation resulting from automatic processing is often too close to natural language (literal semantics), making it necessary to transform it by means of rules in order to make it relevant at the level of the formal design specification. This third assumption only appeared after a first series of experiments. Before its introduction, the proposed formalization was quite different from that produced by the designers themselves. The main reason for this distance was the excessive use of high level thematic relations (agent, object, etc.), which were able to represent the meaning of the sentences but were much less relevant for technical specifications.

- A4: the building of a logical description from an intermediate representation can serve as a pivot for translation to different design specification languages. This assumption is the result of an unanimous finding among designers: a particular language rarely lends itself to modelling the entire design specification. This is at the origin of the multi-specification paradigm [30], which consists of using, in the same design specification, the language that is most conducive to providing a description for a given module. Once the logical description is produced and with the availability of the appropriate translator, it is then possible to translate its content into the target formal design language.

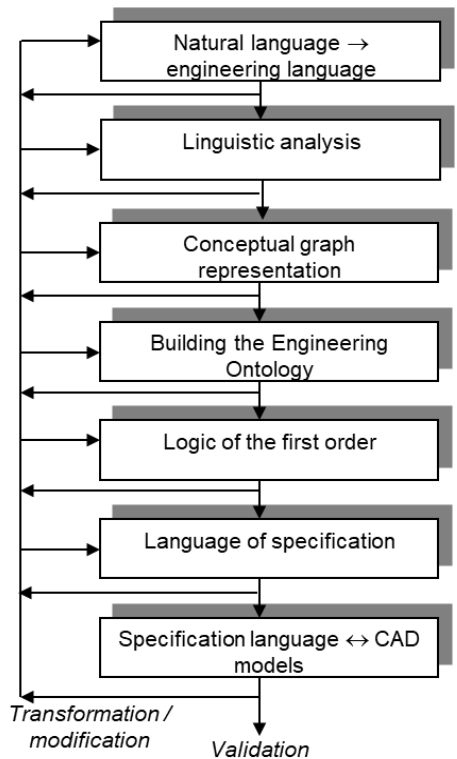

**Figure 1.** Proposed specifications formalization method

According to these four strong assumptions, the proposed method of engineering specifications consists of 7 phases (Figure 1):

- Phase 1: from natural language to the engineering language. This phase performs the extraction of words with the highest density (introduction of fuzziness).
- Phase 2: linguistic analysis. This phase produces three plans: syntactic, semantic, and conceptual.
- Phase 3: conceptual graph representation. This phase formalizes the previous representation in conceptual or actancial graphs.
- Phase 4: building Engineering Ontology. This phase builds ontological representation from conceptual graphs and acquired ontology.
- Phase 5: logic of the first order. This phase transforms conceptual graphs and engineering ontology into expressions of the logic of the first order.
- Phase 6: language of specification. This phase allows passing from the logic of the first order to the specification.
- Phase 7: chaining specification language toward *CAD* models. This is the final phase.

## 2.1. From natural language to the engineering language

Taking into account the static/kinematic/dynamic cognitive trichotomy, understanding a design specification text is identifying and then translating three elements: 1) the static situations of spatio-temporal location and characterization of objects; 2) the kinematic situations of moves in a space or of state changes attributed to objects; and 3) dynamic situations of moves or changes of states caused by an external force. There are, thus, three levels of representation: 1) the *cognitive*, generated from cognitive archetypes; 2) the *conceptual*, organized into predicates and arguments; and 3) the *linguistic*, organized from grammatical schemas specific to a language [31]. To state a description is: 1) to integrate cognitive archetypes into predicative conceptual schemes (predicative structures: predicates, arguments and/or actants) and 2) to encode these predicative structures into specific linguistic systems in the form of grammatical schemas specific to the language used (i.e., the lexical category arrangements such as nouns, verbs, prepositions, adjectives, adverbs, etc. (Figure 2). This enunciation scheme is very close to Vygotski's [27].

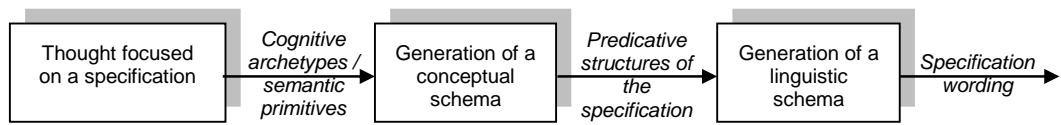

**Figure 2.** Reference scheme for specification wording

During this phase, the designers identify the components of the product they specify (Figure 3, as an example [48]), and the functions of this product (Figure 4, as an example [48]).

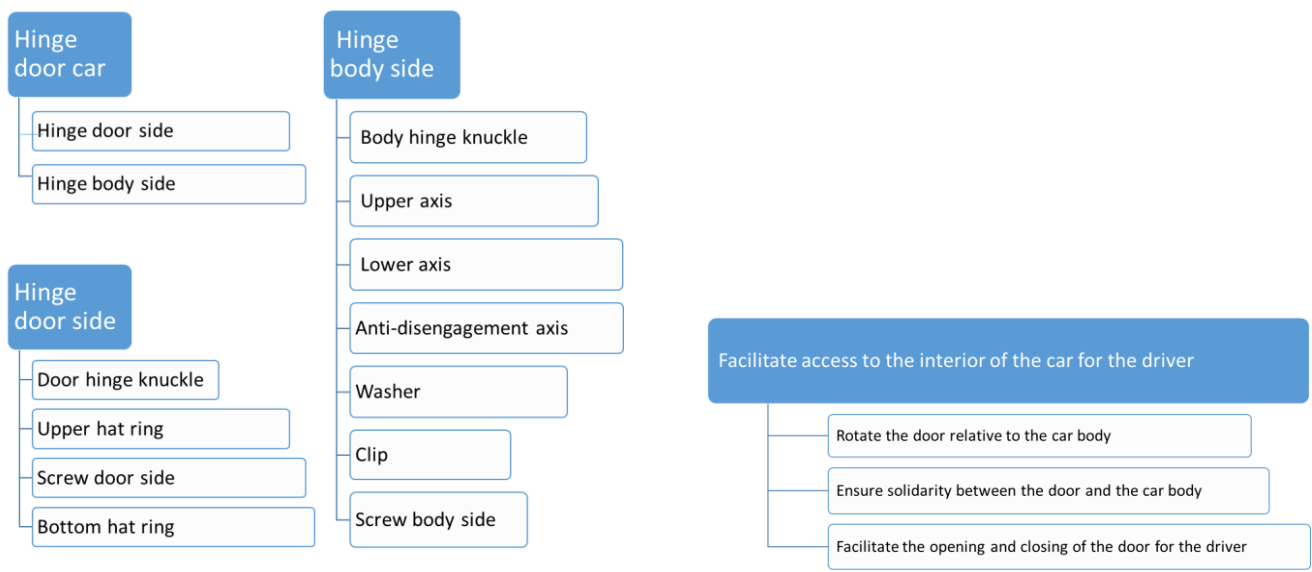

**Figure 3.** List of components of a product as an example

**Figure 4.** List of functions of a product as an example

## 2.2. Linguistic analysis

According to Chomsky [32], language has three fundamental properties: it is intentional (someone has the intention to transmit something to someone else), syntactic (it has the organization, structure and coherence of a speech act), and enunciative (it transmits information). Although the ambition of *NLP* is to propose methods and tools that cover all three properties, the most convincing results are focused on the last two.

It is usual in computational linguistics to consider five levels for the written language: morphology, lexicon, syntax, semantics, and pragmatics. These level are interdependent but, more importantly for this process, stratified. The stratification of written language allows a clear presentation of the elements of the natural language.

- Morphology allows the recognition of basic textual units or words. Computational morphology must, then, propose procedures for the identification of words (or segmentation) and the identification of terms according to their canonical form (or lemmatization). In the analysis of written texts, the word is the first linguistic unit that can be accessed easily, according to the formal criteria of writing. Subsequently, individual words can be broken down into smaller significant units or, contrarily, combined into larger significant units (phrases). The lexical analysis breaks down into two operations: segmentation and morphological analysis.

- Syntax concerns the study of the organization of words into groups of words and groups of words into sentences. Syntactic organization can be easily translated into a syntagmatic structure or a dependency graph. The strategies of parsing syntax, more or less formal, are very numerous: analyses in constituents, formal grammars, grammatical labelling probabilistic, etc. This work relies on two particular formalisms: The Context Free Grammar (*CFG*) [33] and the Lexical Functional Grammar (*LFG*) [34]. The *LFG* formalism has three levels: (1) the c-structure (constituent analysis), described by rules of productions of an out-context grammar, represents the syntactic structures; (2) the f-structure (functional description), composed of "function-value" pairs, highlights the grammatical functions (subject, object, etc.) - a function, named *Pred*, maps the syntactic functions and semantic roles of a predicate; and (3) s-structure (semantic structure), a semantic projection from the c-structure that allows retention of only the predicative structure.

- Semantics concerns the study of the meaning of words and sentences. It studies them by establishing the relations that exist between the sentences, on the one hand, and the objects, relations, actions, actors, places, modalities and events of the world, on the other. But to produce a good representation of the sentence also presupposes the ability to deal with the semantic flexibility (i.e., ambiguities, paraphrases, references, etc.) present in all natural language. There are two approaches to this challenge: 1) lexical semantics, based on primitives and relations between elements, such as Fillmore's case grammars [35], Schank's action primitives [36], Desclés's cognitive archetypes [31], Pottier's general semantic (Figure 5) [37], *STDP* (Stanford Typed Dependency Parser) [38], notions of prototypes and mental models; and 2) grammatical semantics, such as Montaguë's semantics, *DRT* [39]. As for the relations between semantics and automatic treatments, they are amply described in [40].

- Pragmatics is concerned with the effects of meaning in context (the situation of communication, for example), which results in a relation between a sentence *P*, a speaker *L*, a situation of enunciation *S*, and an interpretation *I*. Among the theories and proposed models, speech / dialogue acts, scripts, and plans are often used [41].

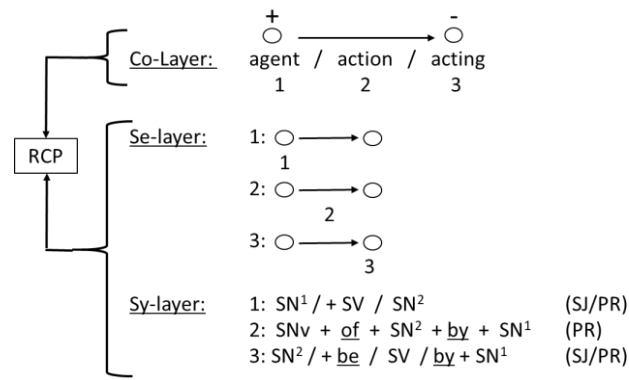

**Figure 5.** From cognitive plan (*Co-layer*) to semantic plan (*Se-layer*) and syntactic plan (*Sy-layer*).

For example, a simple set of grammar rules can be used to analyse a sentence such as: "Access to the interior of the car for the driver" (1-4):

$$s:- \quad vp + np \quad (1)$$

$$vp:- \quad v \quad (2)$$

$$np:- \quad np* \quad (3)$$

$$np:- \quad p + d + n \quad (4)$$

where *s* is a sentence, *vp* is a verb phrase, *np* is a noun phrase, *v* is a verb, *n* is a noun, *p* is a preposition, and *d* is a determiner.

The result of the analysis with these grammar rules is given in the following figure as a syntactic tree (Figure 6).

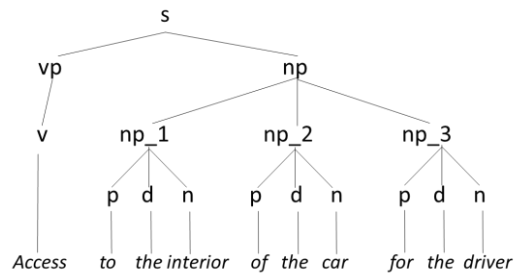

**Figure 6.** A syntactic tree as an example.

## 2.3. Conceptual graph representation

The fundamental problem of any representation of knowledge is the proposition of a sufficiently precise and formal notation or representational framework allowing this representation. In the conceptual graphs (*CG*) model, elementary objects are concepts and relations, which is the formalism of the family of semantic networks. Each proposition or fact is represented by a *CG* which is constructed by directed arcs, connecting concepts and relations (see Figure 7, for example). Rules offering the possibility of joining or dissociating *CGs* (joints and projections) are given. A formal correspondence (isomorphism) with the logic of the first-order predicates is established for a basic kernel. Thus, the formalism of the *CG* is both a "psychological" model in its form, giving it great readability, and a system of mathematical proofs with solid axiomatic foundations, making it formal.

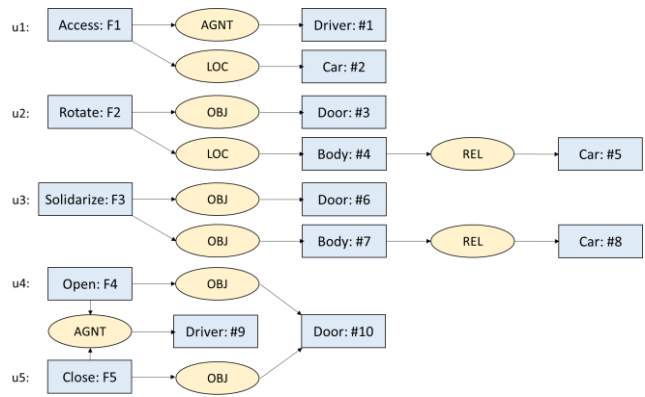

**Figure 7.** Conceptual graphs as examples.

The transition from linguistic analysis to the representation of extracted knowledge, then, consists in the semantic translation of the syntactic structure in the form of *CG* (see Figure 8). For this, case grammars determine the different thematic roles filled by the constituents of a sentence using information acquired on the order of words, prepositions, verbs, and context. In other words, the parser determines how the noun phrases of a sentence are related to verbs - the semantic role that specifies how an object participates in the description of an action. The sources of information activated for the purposes of this analysis are the semantic lexicon and the *LFG* description.

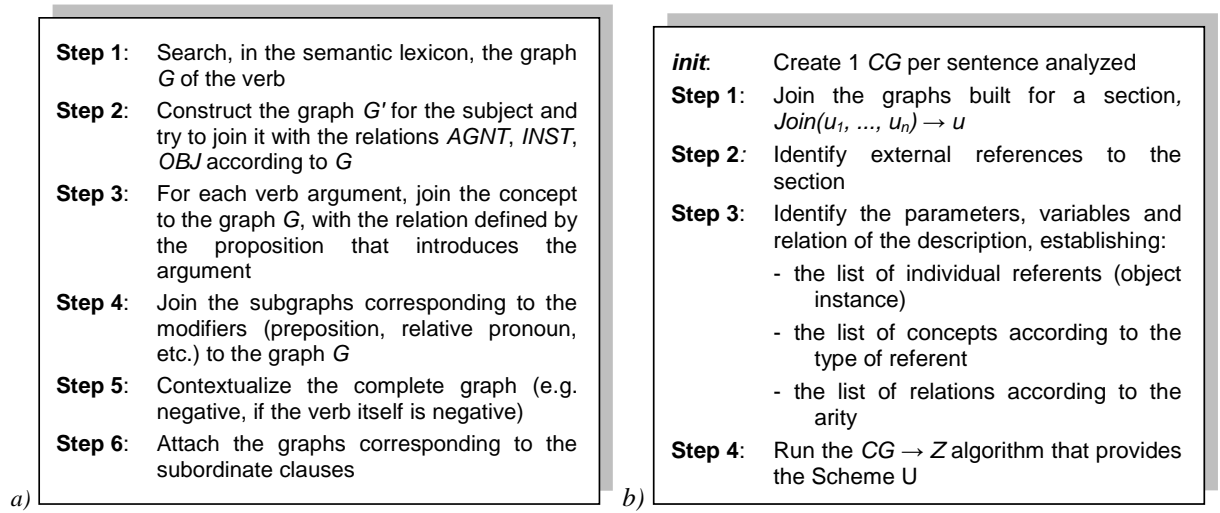

*a)* *b)*

**Figure 8.** a) *CG* building from the functional structure, and b) building of a formal description from the *CGs* of a specification

## 2.4. Building the Engineering Ontology

This engineering ontology construction step consists of extracting the lexical information contained in the specifications and determining the privileged links between the words [42]. It can be performed in two ways:

- Automatically from texts. A co-occurrence study of words, based on lexical proximity analysis, also identifies compound words, phrases, predicative relationships, and domain-specific sentence schemes. The joint use of statistical filtering techniques, such as mutual information [43], makes it possible to increase the relevance of the results obtained. *NLP* tools have emerged to help the building of terminologies and to acquire ontologies from corpora.

- Semi-automatically (as in this work). The approach of acquiring knowledge consists in extracting from a digital dictionary the definitions of terms previously recognized as concepts and, then, in order to describe them in a semantic dictionary, presenting them in the form of *CG* [44], for example. The goal of the treatment is to increase the knowledge base of the application domain by integrating semantic information from a dictionary and expanding the ontology of the model through the definitions of each concept. Once the contents of a definition are analysed, it is possible to build a corresponding *CG* and to include it in a canonical base (See Figure 9 for examples).

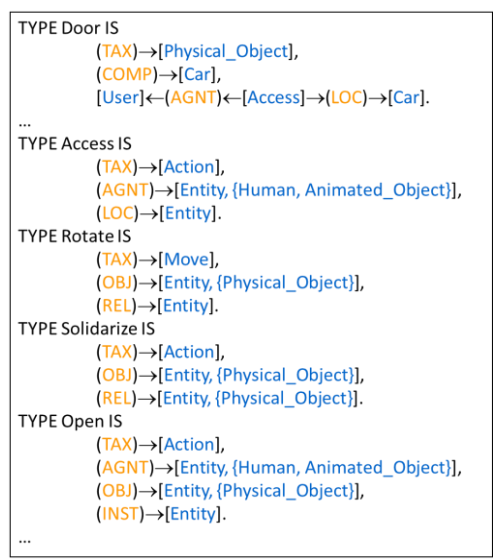

**Figure 9.** *CG*s of components and functions as examples

## 2.5. Logic of the first order and language of specification

Sowa defined the operator $\Phi$ which makes a formula in the predicate logic of the first order correspond to every *CG* [29]. For instance, the *CG u* representing the sentence "Access to the interior of the car" (5), will have for equivalent the formula $\Phi(u)$ (6):

u: [access: *]→(LOC)→[car: #1].                    (5)

$\Phi(u)$: $\exists x, access(x) \wedge LOC(x,\#1) \wedge car(\#1)$        (6)

Through the isomorphism between *CG* and predicate logic, it becomes possible to propose a formal language of specification, as illustrated by the following translation in the *Z* language (or *Object-Z* [45]). This notation has the advantage of making more explicit the representation in logic of the first order. It allows specifying a system/object by describing its state and the operations that modify it. For this, a *Z* specification consists of a sequence of paragraphs with schemas, variables, and basic types. A schema consists of a signature or a collection of typed variables and a property on that signature, called the predicative part. In order to build the formal description that corresponds to an informal specification, the following steps are followed: 1) extract the elements from the formal description; 2) identify and insert the indispensable elements that are not included in the module to be specified; and 3) establish the logical formulas (i.e., the pre-conditions, post-conditions, properties, etc.) that correspond to the elements collected and their links (see Figure 10). The final phase consists of *Z* modelling of *CG*s constructed from the informal specification.

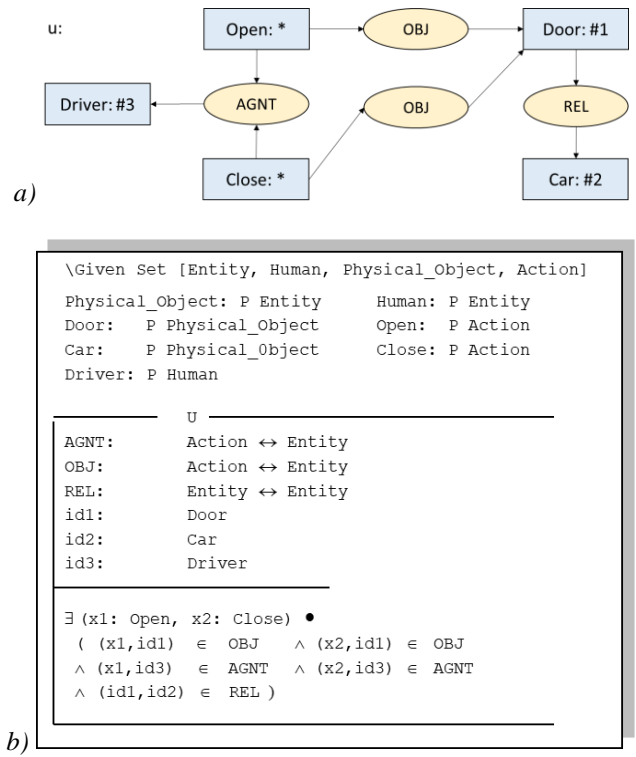

**Figure 10.** a) *CG* corresponding to the specification "A driver must be able to open and close the door", and b) the resulting schema *Z*

## 2.6. Chaining specification language toward *CAD* models

A bill of materials (*CAD-BOM*) represents the structure of the files in *CAD* modelling. Figure 11 shows the *CAD-BOM* of the car door hinge (corresponding to components listed in figure 3).

In the case of an existing product, the outputs (variables, functions, formulas, etc.) of the language of specification are transmitted to each file (part file or assembly file) in *CAD-BOM* by communicating agents (see Figure 12). In the case of a new product, *CAD-BOM* is also created by the outputs (variables, functions, formulas) of the first order logic.

During *CAD* modelling, a formula is evaluated as true or false, given a variable assignment that associates a created geometric or topologic element in *CAD* with each variable: for instance, check rules of the form, *IF condition THEN action*, in Knowledgeware Advisor of Computer Aided Three-Dimensional Interactive Application (*CATIA*) are programmed for the evaluation. Thus, during *CAD* modelling, a *CAD* model is evaluated in terms of its satisfaction of the specifications.

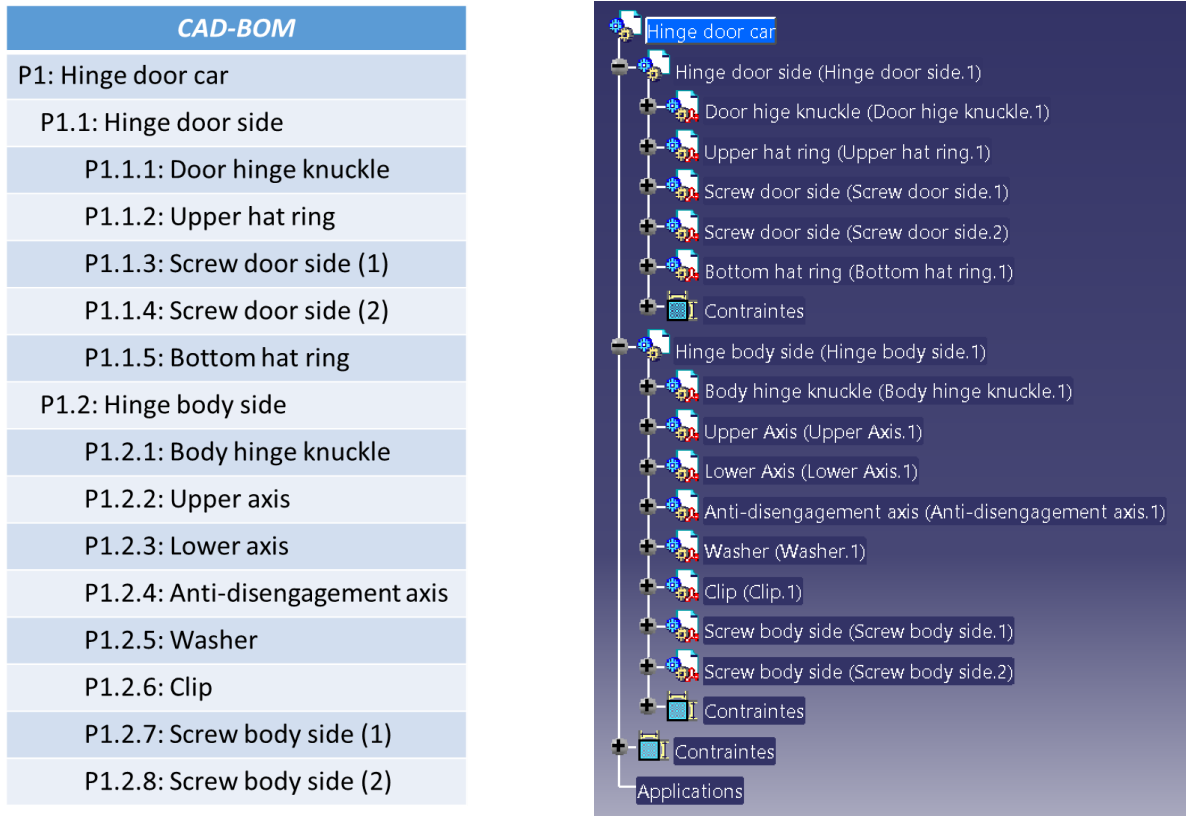

**Figure 11.** *CAD-BOM* of the car door hinge

**Figure 12.** Product structure of the car door hinge in Computer Aided Three-Dimensional Interactive Application (*CATIA V5 R21*)

## 3. APPLICATION AND RESULTS

After having worked on software design according to the *NIAM* method (Nijssen Information Analysis Method) [46], this research proposes an automatic process of diagrammatic representation (close to the UML language) of these descriptions [44]. This process of linguistic analysis incorporates two main phases: acquisition of knowledge of the domain and linguistic analysis, conforming with that exposed previously with the exception of the semantic level. For semantic analysis, it relies on case grammars. The latter make it possible to recognize the different thematic roles filled by the components of the sentence, using the information collected on the order of the words, the prepositions, the verbs, and the context. The parser determines the meaning of the noun groups of a sentence in relation to its verbs; the semantic role can, then, specify how an object participates in the description of an action or a state. This process of linguistic analysis is entirely object oriented.

The syntactic-semantic parser uses a case grammar and translates the resulting representation as a *CG*. This parser is based on a case analysis for two reasons [47]: 1) syntactic and semantic parsing in two separate modules provides an advantage over a computer development point of view and 2) there is an advantage in directly mapping the syntactic form and the semantic cases. This approach defines a set of semantic cases as well as a set of prototypes that characterize the case structure of verbs encountered in design specification corpora. The representation of a sentence is composed of a modality and a proposition. The modality contains information on the negation, the interrogation, the time, and the mode of the verb (the obligation, the possibility, etc.). The proposition results from the identification of the verb and the semantic relations which link the noun groups to the verb. The analysis concludes with semantic representation, integrating the verb and its semantic cases (sometimes prototypical) instantiated by the corresponding conceptual elements. In the context of an agent design approach, the parser effectively distinguishes the agent entities from the object entities.

The implementation of the method outlined in the previous section 2 (Figure 1) is illustrated by the following scenario in which designers discuss the functional requirements of a car door hinge [48] (see Figure 13). While the designers discuss and share their visions of the hinge, an agent-based tool (*EGEON*: **E**nginering desi**G**n s**E**mantics elab**O**ration and applicatio**N**) translates (writes) their specifications or proposals. The semantic representation of the description of the hinge is then built automatically and the different phases of the method are deployed.

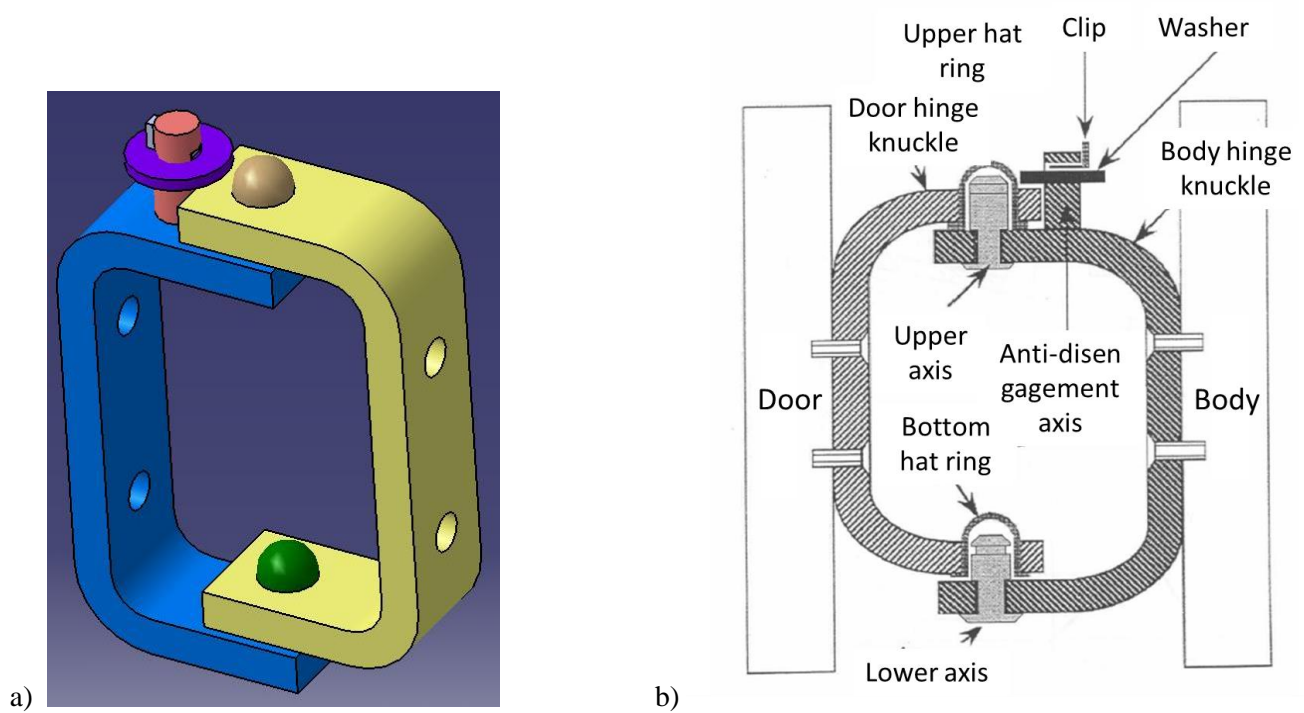

**Figure 13.** A car door hinge

The agent-based tool *EGEON* works on the first 4 phases of the proposed method (Figure 3), although the agentification of the approach of intelligent requirements engineering from natural language is not developed in this article. *EGEON* makes it possible to represent the information transmitted by the designers on two levels:

- In the first level of representation the functions are independent (Figure 14). As many graphs of knowledge are generated as sentences / information transmitted. This is the first phase of understanding the function (and then the transformation of the function into a solution).

- In the second level of representation, a knowledge network is generated by merging the previous graphs (Figure 15). This network shows the interactions between the functions. This representation is also very important because it can inform the designer, for example, about the conflict between functions and their parameters. For instance, the parameters of the actions "rotate" and "solidarize" interacting with "body" and "door" can be conflictual. Similarly, the actions "open" and "close" related to "door" and "driver" are

**11**

semantically conflictual. The network can also assist the designer in explaining the causality and the deductions, and showing the interaction between an established function and a newly defined one.

## 4. DISCUSSION AND CONCLUSION

This paper proposes an intelligent method for requirements engineering from natural language and their chaining toward *CAD* models. The transition from linguistic analysis to the representation of engineering requirements consists of the translation of the syntactic structure into semantic form represented by conceptual graphs. Through this isomorphism between conceptual graphs and predicate logic, a formal language of the specification is proposed which is translated in language Z, which in its turn is chained in Computer Aided Three-Dimensional Interactive Application (*CATIA*) models.

The tool (*EGEON*: **E**nginering desi**G**n s**E**mantics elab**O**ration and applicatio**N**) represents the semantic network of engineering requirements. It can assist the designer to explain the causality and the deductions.

Intelligent requirements engineering from natural language and its chaining toward *CAD* models improve the time and quality of requirements.

The proposed formalization enhances the reliability and the pertinence of the requirements in both *CAD* models and, in general, *PLM* systems. The production of models in time also informs the structure of the design process and its management.

The proposed method can not only be used to develop and improve design but also the basis of the social process of designing through discussion. The proposed method focuses on the team discussion during which design decisions are made. Therefore, natural language processing and engineering language building are important parts of requirements engineering.

From this point of view, intelligent requirements engineering from natural language and its chaining toward *CAD* models assist the process of collaborative working and team building. They make it possible for a design team to use intelligent models in dialectical, exploratory, and open ways.

The proposed intelligent requirements engineering model suggests open-ended situations and invites further contributions. It does not suggest a finished system and does not make a closed *AI* application. Instead, the proposed modelling systems widen the application of *AI* in engineering requirements.

Development of requirements by specialist team members working together enables the development of better design proposals. However, intelligent requirements engineering by the proposed systems should not just represent the proposals but also support positive actions or behaviours in the design environment. For example, an intelligent requirements engineering system should allow an analysis of misrepresentation and biased representation. Such shortcomings can have an undue negative impact on the development of a design as well as on the social process of designing. Thus, the future scenarios for the development and the use of the tool *EGEON* concern the assertion of requirements and making recommendations.

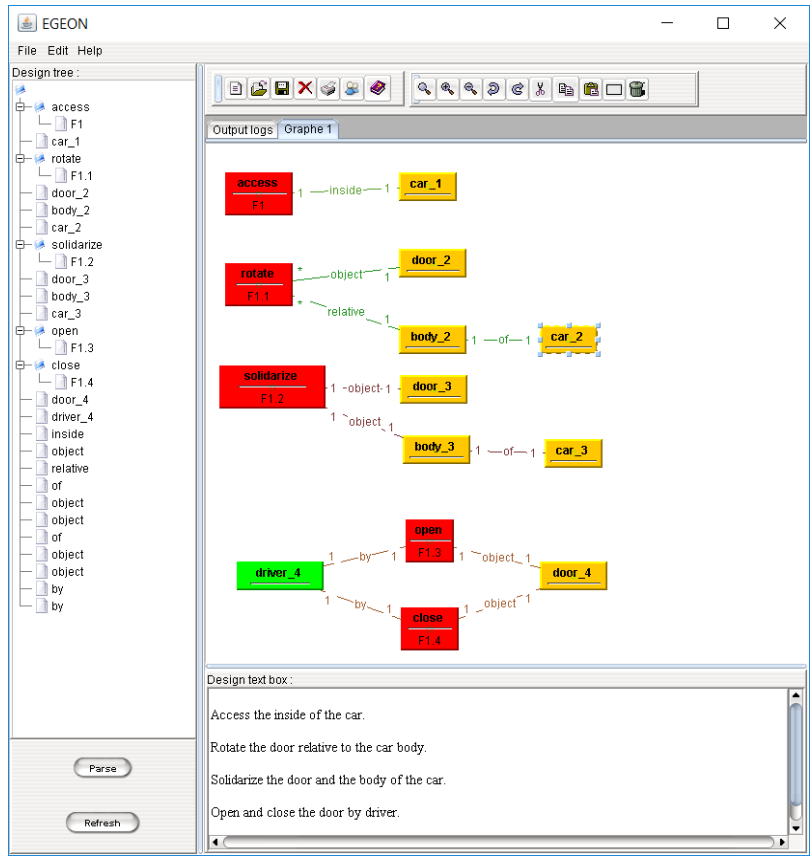

**Figure 14.** First level of semantic representation in the *EGEON* environment.

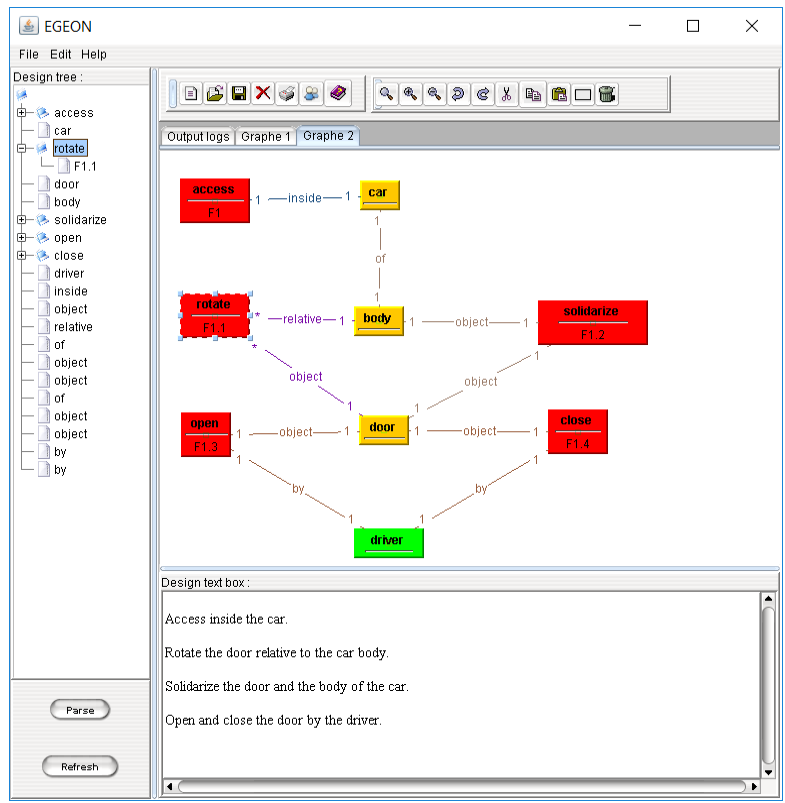

**Figure 15.** Second level of semantic representation in the *EGEON* environment.